\documentclass{article}

\usepackage{PRIMEarxiv}
\usepackage[nohyperlinks, nolist]{acronym}

\usepackage[utf8]{inputenc} 
\usepackage[T1]{fontenc}    
\usepackage{hyperref}       
\usepackage{url}            
\usepackage{booktabs}       
\usepackage{amsfonts}       
\usepackage{nicefrac}       
\usepackage{microtype}      
\usepackage{lipsum}
\usepackage{fancyhdr}       
\usepackage{graphicx}       
\graphicspath{{media/}}     

\title{Detection Transformers Under the Knife:\\ A Neuroscience-Inspired Approach to Ablations
}

\author{
\href{https://orcid.org/0000-0002-4179-6733}{\includegraphics[scale=0.06]{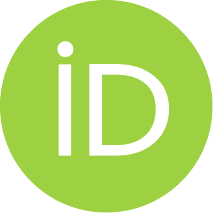}\hspace{1mm}Nils Hütten}\thanks{Corresponding author. {\it E-mail address:} Nhuetten@Uni-Wuppertal.de}, \href{https://orcid.org/0009-0002-5191-901X}{\includegraphics[scale=0.06]{images/orcid.pdf}\hspace{1mm}Florian Hölken}, \href{https://orcid.org/0000-0003-0080-6285}{\includegraphics[scale=0.06]{images/orcid.pdf}\hspace{1mm}Hasan Tercan},
\href{https://orcid.org/0000-0002-1969-559X}{\includegraphics[scale=0.06]{images/orcid.pdf}\hspace{1mm}Tobias Meisen} \\
 Institute for Technologies and Management of Digital Transformation (TMDT) \\
 University of Wuppertal\\
 Rainer Gruenter Str. 21, 42119, Wuppertal, Germany \\
}

\begin{document}

\begin{acronym}[DDETR]
\acro{cnn}[CNN]{convolutional neural network}
\acro{mhsa}[MHSA]{Multi-head self-attention}
\acro{mhca}[MHCA]{Multi-head cross-attention}
\acro{xai}[XAI]{explainable artificial intelligence}
\acro{detr}[DETR]{detection transformer}
\acro{ddetr}[DDETR]{deformable detection transformer}
\acro{dino}[DINO]{DETR with improved denoising anchor boxes}
\acro{giou}[gIoU]{generalized intersection over union}
\acro{iou}[IoU]{intersection over union}
\acro{mgiou}[mgIoU]{mean generalized intersection over union}
\acro{vt}[VT]{vision transformer}
\acro{swin}[Swin]{shifted window transformer}
\acro{gpt}[GPT]{generative pretrained transformer}
\acro{bert}[BERT]{Bidirectional encoder representations from transformers}
\acro{llama}[LLaMA]{Large Language Model Meta AI}
\acro{mlp}[MLP]{multi-layer perceptron}
\acro{gradcam}[Grad-CAM]{gradient weighted class activation mappings}
\acro{ann}[ANN]{artificial neural network}
\acro{nlp}[NLP]{natural language processing}
\acro{coco}[COCO]{common objects in context}
\acro{ffn}[FFN]{feed forward networks}
\acro{map}[mAP]{mean average precision}
\acro{ann}[ANN]{artificial neural network}
\acro{nia}[NIA]{neuroscientifically inspired ablation}
\acro{qe}[QE]{query embedding}
\end{acronym}

\maketitle

\begin{abstract}
    In recent years, \ac{xai} has gained traction as an approach to enhancing model interpretability and transparency, particularly in complex models such as detection transformers. Despite rapid advancements, a substantial research gap remains in understanding the distinct roles of internal components -  knowledge that is essential for improving transparency and efficiency. Inspired by neuroscientific ablation studies, which investigate the functions of brain regions through selective impairment, we systematically analyze the impact of ablating key components in three state-of-the-art detection transformer models: \ac{detr}, \ac{ddetr}, and \ac{dino}. The ablations target \acp{qe}, encoder and decoder \ac{mhsa} as well as decoder \ac{mhca} layers. We evaluate the consequences of these ablations on the performance metrics \ac{giou} and F1-score, quantifying effects on both the classification and regression sub-tasks on the COCO dataset. To facilitate reproducibility and future research, we publicly release the \href{https://github.com/tidujiie/DeepDissect}{DeepDissect} library.
    
    Our findings reveal model-specific resilience patterns: while \ac{detr} is particularly sensitive to ablations in encoder \ac{mhsa} and decoder \ac{mhca}, \ac{ddetr}’s multi-scale deformable attention enhances robustness, and \ac{dino} exhibits the greatest resilience due to its look-forward twice update rule, which helps distributing knowledge across blocks. These insights also expose structural redundancies, particularly in \ac{ddetr}'s and \ac{dino}'s decoder \ac{mhca} layers, highlighting opportunities for model simplification without sacrificing performance.
    This study advances \ac{xai} for detection transformers by clarifying the contributions of internal components to model performance, offering insights to optimize and improve transparency and efficiency in critical applications.
\end{abstract}


\section{Introduction}
    \label{sec:intro}
    The advent of deep learning has revolutionized computer vision, with \acp{cnn} advancing tasks like classification, object detection, and segmentation. More recently, \acp{vt} have emerged, leveraging attention mechanisms to handle long-range dependencies in image data \cite{Vaswani.2017, Dosovitskiy.2020}. Originally developed for sequence tasks, transformers now rival or outperform CNNs in visual benchmarks \cite{liu2021swin,Yang.01.07.2021}, driving rapid adoption in both academia and industry \cite{Khan.2021, Hutten.2024}.
    The pursuit of benchmark performance and inference times, drives increasing model complexity, as seen in the evolution of models such as BERT \cite{Devlin.2018}, GPT \cite{Brown.2020}, LLAMA, and \ac{vt} models like ViT, Swin, or Focal Transformer \cite{Yang.01.07.2021} \cite{Dosovitskiy.2020,liu2022swin,Yuan.22.11.2021,Yang.22.03.2022,Wang.10.11.2022}. Despite their success, these models pose significant challenges regarding explainability and interpretability. Although \ac{xai} is an active field, specific investigations into \acp{vt} remain limited \cite{Stassin2024Explainability, Holistic2023Explainability, Explainable2023Review}. In particular, methodology to reveal how knowledge acquired during training is internally represented remains underexplored \cite{Khan.2021}, yet is crucial for building transparent and trustworthy models, especially in high-stakes domains like autonomous driving and medical diagnosis.
    
    The field of neuroscience offers valuable inspiration for addressing this gap. Techniques like ablative brain surgery—selective removal of neural tissue—have been instrumental in identifying brain regions responsible for specific cognitive and motor functions \cite{Scoville.1957, Rolls.2000, Ungerleider.1982, LeDoux.1996}. Inspired by this, early studies by Meyes et al. \cite{Meyes2019} applied \acp{nia} to simple \acp{mlp} and \acp{cnn} to enhance interpretability.
    
    In our work, we extend this approach to modern \acp{vt} to address key challenges in \ac{xai} for object detection. Traditional \ac{xai} methods - such as feature attribution and gradient-based techniques - primarily focus on identifying influential input regions, yet often fail to provide insights into the internal structure and functional roles of model components. In contrast, \ac{nia} studies offer a systematic way to isolate and analyze their contributions individually, shedding light on how knowledge is represented and utilized within the network. We adapt this methodology to address our central research question "How do architectural developments in detection transformers affect the internal organization of learned representations with respect to the regression and classification subtask?".
    Our contributions are twofold. First, we provide new insights into the knowledge distribution and the performance contributions of key architectural components by conducting \ac{nia} studies on three key VT models for object detection. Second, we release the Python-based library \href{https://github.com/tidujiie/DeepDissect}{\textit{DeepDissect}}, which facilitates the execution of \ac{nia} studies on models within the MMDetection \cite{mmdetection} framework.
    
    Using DeepDissect we carried out a systematic ablation study across three architectures from the detection transformer family, focusing on key model components like ac{qe}s and encoder/decoder projection matrices. For each component, we performed full layer ablations, from 5-50\%, as well as block-wise ablation with a constant ablation level of 30\%. The models were evaluated on the \ac{coco} validation set \cite{Lin.01.05.2014} using both classification and regression metrics to quantify the impact of the ablations on performance. Our results provide a deeper understanding of how knowledge is represented within \ac{detr} models, offering insights into the internal semantics of their components and how each contributes to the models’ overall functionality. By identifying the contribution and actual function of each component, we pave the way for optimizing these architectures, for more efficiency and interpretability.

\section{Related Work}
    The term \emph{ablation} in the context of technical systems was originally introduced  by Newell \cite{Newell.1974} as a method for analyzing the contribution of individual components by removing them and observing the resulting changes in system performance. He argued that, in complex systems composed of modular knowledge sources, ablation could reveal valuable insights about the internal structure and function. Even in deliberately engineered systems, the specific contributions of individual parts are not always well understood, making ablation a pragmatic tool for analyzing internal structure and functional contributions. This principle remains relevant for modern \acp{ann}, where internal mechanisms often remain opaque despite careful design.
    
    In deep learning, ablation studies have evolved into a widely used but broadly defined methodological category. The term is often applied to a wide range of experiments aimed at demonstrating improvements in performance or efficiency. These include hyperparameter tuning (e.g., ResNeXt, Faster R-CNN, FPN, and DETR \cite{Xie.2016,Ren.04.06.2015,Lin.2016,Carion.26.05.2020}), structural modifications such as the omission or replacement of components (e.g., RCNN, Faster R-CNN and DETR \cite{Girshick.11.11.2013,Ren.04.06.2015,Carion.26.05.2020}), and interventions on information pathways (notably in FPN and DETR \cite{Lin.2016,Carion.26.05.2020}). Additionally, studies like Hameed et al. \cite{hameed2022.input-ablations} examine input-level ablations through image perturbation. These experiments are typically conducted pre-training, but also rarely are done post-training. Further, while all these approaches are commonly labeled as ablation studies, they are often conducted to retrospectively justify design choices—demonstrating that a newly introduced module or mechanism yields a measurable benefit, rather than to gain insight into the model’s internal structure per se.
    
    However, an alternative perspective, inspired by neuroscience regards ablations as a means for improving model interpretability rather than solely measuring performance. The concept of \ac{nia} involves disabling trainable weights by setting them to zero in a trained \ac{ann} to block signal flow.
    Pioneering work in this direction by Meyes et al. \cite{Meyes2019} examined the performance of a \ac{mlp} and a VGG-19 \ac{cnn} trained on the classification datasets MNIST and ImageNet, respectively. Their findings revealed that certain \ac{mlp} units or \ac{cnn} layers play a disproportionately substantial role in determining the overall outcome, either by representing general concepts or class-specific features. These insights were generalized by Vishnusai et al. \cite{Vishnusai.2020} through \ac{nia} studies with different models trained on MNIST. Lilian et al. \cite{Lilian2019} extended the approach to robot control based on reinforcement learning, demonstrating that \ac{ann} exhibit a degree of logical organization, in addition to resilience to structural damage.
    
    Furthermore, the blackbox nature of \ac{ann} can be further addressed by combining ablations with methods such as \ac{gradcam} \cite{GradCAM.2017}. For example, the transparency and interpretability of a one-dimensional \ac{cnn} used for failure prediction in industrial sensor data were improved by demonstrating that its learned filters aligned with \ac{gradcam} explanations and were intuitively understandable to domain experts \cite{Transparent2021}. While methods like \ac{gradcam} highlight relevant input regions by mapping model decisions back to input features, ablation studies offer a complementary perspective: they probe the internal structure of the model by selectively disabling components, thereby revealing how specific internal representations contribute to overall function.
    
    While the model compression technique \emph{pruning} often employs similar experimental setups, their underlying motivation fundamentally differs from that of \ac{nia} studies. Recent applications of pruning enabled \ac{nlp} models like SparseGPT \cite{Frantar2023SparseGPT}, ShortGPT \cite{Men2024ShortGPT}, Wanda \cite{Sun2024Wanda}, UPop \cite{Shi2023UPop} and Shortened LLaMa \cite{Kim2024ShortenedLLaMA}. There have also been pruning endeavours on \ac{vt} models like \ac{swin}, ViT \cite{FangYu.2022,LuYu.2023,YifeiLiu.2024} and also \ac{detr} models \cite{Su.2025,Sun.2024}.
    The primary goal of pruning is to reduce model size and computational cost, and it usually does not distinguish between pruned parameters at the component level. Furthermore, most methods introduce additional learnable components to determine which (partial) network structures are to be excluded, making them unsuitable for \ac{xai} purposes. Hence, due to the fundamentally different objectives and level of detail of assessments, meaningful comparisons between \ac{nia} studies and pruning are not adequate.

\section{Methodology}
    Guided by our aforementioned key research question, we conduct a structured \ac{nia} study across three key \ac{detr}-based architectures, all trained on the \ac{coco} dataset. These models represent significant advancements in detection transformer technology: the original \ac{detr} \cite{Carion.26.05.2020}, \ac{ddetr} \cite{DDETR.Zhu.2021}, and \ac{dino} \cite{zhang2023dino}.
    \ac{ddetr} introduced two new architectural features: deformable attention layers and multi-scale feature processing. Deformable attention reduces computational cost by attending only to a set of $k$ predicted offsets with regard to a reference point per attention head. These reference points are an additional input to the first decoder block and are defined and optimized the same way as \acp{qe}. The multiscale features are extracted from different layers in the ResNet-50 backbone and jointly used as the input for the encoder \ac{mhsa}. Although \ac{ddetr} also explores bounding box refinement via dynamic priors and two-stage architectures, we exclude these variants to keep the scope of the study manageable and to maintain a certain architectural difference to DINO.
    \ac{dino} introduced three novel concepts: contrastive denoising training to reduce duplicate detections; mixed query selection fusing encoder outputs with \acp{qe}; and the look-forward twice update rule to achieve faster convergence through an additional gradient path in neighboring decoder blocks.
    
    To systematically investigate the models, we designed a set of coherent ablation studies grounded in the base \ac{detr} architecture for consistency. In this regard, it seemed logical to focus on the components that distinguish \ac{detr} from \acp{cnn} and \ac{nlp} transformers: \ac{mhsa}/\ac{mhca} layers, learnable \acp{qe} forming the decoder input, and additional auxiliary loss terms to stabilize training.
    
\begin{figure}[h!]
      \centering
       \includegraphics[width=0.45\linewidth]{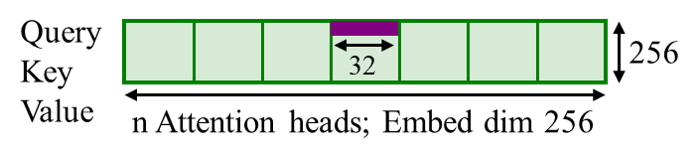}
       \caption{Visualization of multi-head attention ablations on $1x32$ subunits representing attention heads in projection matrices.}
       \label{fig:projection_matrix_ablation}
    \end{figure}

    We decided to ablate the input projection matrices (\begin{math}W_{q}, W_{k}, W_{v}\end{math}) of the encoder \ac{mhsa} , decoder \ac{mhsa}, and decoder \ac{mhca} since these directly impact the attention computation. The first set of ablation studies is conducted simultaneously across all transformer blocks, removing 5\%, 15\%, 30\%, and 50\% of the weights, referred to as \emph{full ablations}. Additional \emph{block-wise ablations} consider the distribution of learned representations across transformer blocks, targeting the desired component in one block at a time. They were conducted only on the attention layers at 30\%, due to the significance of the effect in full ablations and resource constraints. We perform ablations by randomly zeroing the specified percentage of weights within the targeted component, effectively blocking signal flow. We then evaluate the model on the validation set, comparing its performance against the unmodified state (see Fig. \ref{fig:projection_matrix_ablation}). Each experiment includes 100 random configurations per ablation level, chosen to balance statistical robustness and computational efficiency (see Appendix Fig. \ref{fig:query_embedding_std_variation} for variance stabilization trends).

    \acp{qe} are ablated individually, at the scalar level. For the multi-head attention components we ablate $1x32$ vectors from projection matrices, preserving the coherence of subunits within the attention mechanism (Fig. \ref{fig:projection_matrix_ablation}, right).
    To discern whether the classification and regression aspects of the object detection task are organized within specific model regions or evenly distributed, we apply two distinct metrics and bounding box matching approaches. For classification performance, we use the F1-score, while regression performance is evaluated with the \ac{giou} \cite{gIoU.paper}.
    To calculate the F1-score, we first match the predicted bounding boxes with the corresponding ground truth boxes based on the \ac{giou} via the Hungarian algorithm. After excluding unmatched or invalid detections, the scores are determined using a weighted average across classes, which considers the varying instance counts per class, yielding an overall metric that accurately reflects the model’s classification performance across the entire dataset.
    
    The \ac{giou} adjusts the \ac{iou} by incorporating the area of the smallest enclosing box containing prediction and ground truth, which quantifies regression accuracy even for bounding boxes that do not overlap with the ground truth, offering a more comprehensive assessment of spatial alignment. This adjustment penalizes predictions that are distant from the ground truth, providing a more robust measure of localization accuracy. To obtain an overall metric, we compute the weighted average across classes, giving us the \ac{mgiou}.

\section{Results \& Analysis}
    \label{sec:results}
    In this section, we formulate hypotheses based on the original publications introducing the respective models and examine them in light of our ablation study results. As baselines, we use models trained on the  \ac{coco} dataset, provided by MMDetection \cite{mmdetection}. Their performances are summarized in tab. \ref{tab:initial_performance}. The baseline results highlight the progress in regression performance across the three models - from \ac{detr} to \ac{ddetr} and finally \ac{dino}. Interestingly, classification performances remains largely stable across models, with \ac{detr} and \ac{ddetr} even slightly outperforming the more recent \ac{dino} model in this sub-task.

    \begin{table}[h!]
    \centering
    \begin{tabular}{|l|c|c|}
    \hline
    \textbf{Model}                   & \textbf{\ac{mgiou}\ (\%)} & \textbf{F1-score (\%)} \\ \hline
    \textbf{\ac{detr}}  & 70.35                                     & 85.17                  \\
    \textbf{\ac{ddetr}} & 73.99                                     & 85.67                  \\
    \textbf{\ac{dino}}  & 81.27                                     & 84.88                  \\ \hline
    \end{tabular}
    \caption{Baseline performance of models in terms of \ac{mgiou} and F1-score.}
    \label{tab:initial_performance}
    \end{table}

    \subsection{Detection Transformer Model (\ac{detr})}
\begin{figure}[h]
          \centering
                 \includegraphics[width=1\linewidth]{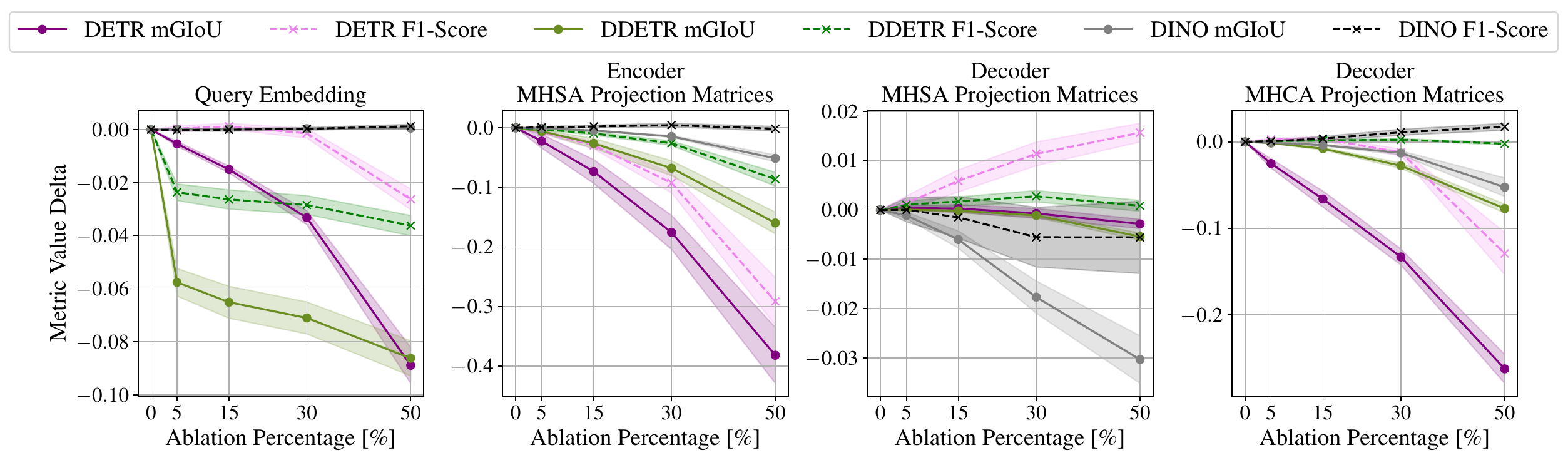}
            \caption{\ac{detr}, \ac{ddetr} and \ac{dino} performance differences for increasing ablation percentages in \acp{qe}, encoder MHSA, decoder MHSA and decoder MHCA. The shaded areas correspond to the standard deviation. Separate visualizations for each model in Appendix Fig. \ref{fig:detr_component_ablations}, \ref{fig:ddetr_component_ablations}, \ref{fig:dino_component_ablations}}
            \label{fig:components_overall}
        \end{figure}

\begin{figure}[h!]
          \centering
           \includegraphics[width=1\linewidth]{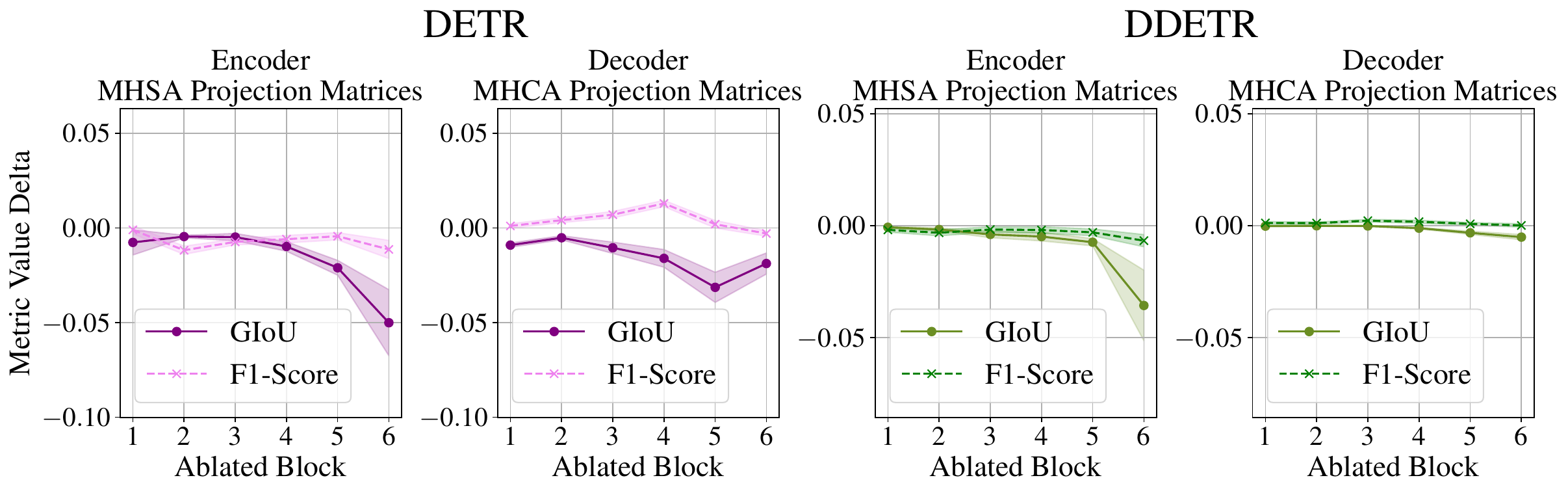}
           \caption{\ac{detr} and \ac{ddetr} performance difference for 30\% block-wise ablations in encoder MHSA and decoder MHCA. The shaded areas correspond to the standard deviation.}
           \label{fig:detr_layers}
        \end{figure}

        The initial DETR paper presented ablation experiments that are typical for the introduction of new models. These included variations in the number of encoder and decoder blocks, removal of \ac{ffn}, and different configurations for spatial positional encodings. Removing all encoder blocks resulted in a performance drop of 3.9 \%p in \ac{map}, while using predictions from the first decoder block instead of the last led to a more substantial reduction of 8.2\%p. The newly introduced \acp{qe} were considered indispensable by the authors, so only the spatial position encodings were removed or their entrypoint into the model was changed. This led to a maximum reduction in \ac{map} of 7.8\%p upon removal. In contrast to our experiments, these modifications were applied prior to training, but still can serve as a sound basis for forming our hypotheses. Based on these results, we expect a continuous decline of \ac{mgiou} and F1-score for the full ablations with increasing ablation percentage, with decoder attention being most sensitive, followed by \acp{qe} and finally encoder \ac{mhsa}. We cannot derive any assumptions from the initial experiments by Carion et al. delineating classification from regression performance, but based on the characteristics of the metrics, regression should be more susceptible than classification, given its direct connection to localization errors. For the block-wise ablations, we expect a continuously increasing performance reduction the deeper the ablated block. In the case of the decoder, this would be consistent with the original paper. For the encoder, it aligns with the widely accepted notion that features become increasingly refined—and therefore more critical to function—the deeper they are processed within the network.
        Figure \ref{fig:components_overall} illustrates the performance differences resulting from the \acp{nia}, with gIoU shown as dot markers on solid lines and F1-score as cross markers on dashed lines. The x-axis indicates ablation percentages, and each plot is labeled with the corresponding model component.
        Contrary to our hypothesis, ablating the decoder MHSA had minimal impact on regression and even slightly improved classification at higher ablation levels. It appears that the re-projection of the queries is not required in the fully trained state, as the information provided to the decoder MHCA layer by residual connections is sufficient. \acp{nia} of the decoder MHCA resulted in the second highest reduction of gIoU and F1-score, and showed a steady decline in regression performance, with a loss of 26\%p \ac{mgiou} at 50\% ablation, while classification performance remained stable until the highest ablation level. This highlights the decoder \ac{mhca}'s central role in integrating encoder features over \ac{mhsa}'s query re-projection function. Although \ac{qe} ablations resulted in significant regression deterioration, \ac{mgiou} only dropped by a maximum of 9\%p, contradicting our hypothesis regarding the ranking of model components. The F1-score was less affected with a maximum decrease of \begin{math}\approx\end{math} 5\%p, supporting their primary function lies in localization as intended by Carion et al. \cite{Carion.26.05.2020}. Encoder \ac{mhsa} ablations led to the highest performance drop—39\%p in \ac{mgiou} and 29\%p in F1-score at 50\% ablation—contradicting our hypothesis based on the \ac{detr} paper by underscoring its essential role in constructing features for both localization and classification. 
        Figure \ref{fig:detr_layers} visualizes the results of the block-wise ablations, following the same concept as Fig. \ref{fig:components_overall}. With the exception that the x-axis depicts the transformer block in which the 30\% ablations were performed.
        We exclude the results of the decoder \ac{mhsa} due to negligible effects. \acp{nia} in the encoder \ac{mhsa} reveal a steadily decreasing \ac{mgiou} from block four onward, indicating that substantial feature refinement mostly occurs in the deeper half of the encoder. Classification displays no clear trend, with F1-score fluctuations between 0.44\%p and 1.1\%p, suggesting that class-discriminative information is redundantly encoded across multiple blocks. Decoder \ac{mhca} ablations, unexpectedly, improved classification across all blocks, peaking at block four with +1.3\%p F1-score. Regression performance declined steadily up to block five but rebounded at block six, deviating from both our hypothesis and the original DETR findings. This implies that only the final three to four decoder blocks meaningfully contribute to regression, while classification ability may be constrained by training dynamics.

    \subsection{Deformable Detection Transformer Model (\ac{ddetr})}
        Our hypotheses for the \ac{ddetr} ablations stem from its two core innovations: multi-scale features and deformable attention. We expect that adopting multi-scale features leads to larger performance losses in both considered metrics, because each ablated value in the projection matrices influences more intermediate results, compared to \ac{detr}. This effect is caused by concatenating the 6.5 larger input along the embedding dimension, before multiplying it with the projection matrices. In contrast, deformable attention likely increases resilience to ablation by reducing the likelihood of ablating critical parameters, through reducing the number of attended tokens per head from $H/32 \times W/32$ (height/width of the input image) to four (single-scale) or sixteen (multi-scale). Given that both the encoder \ac{mhsa} and decoder \ac{mhca} utilize this mechanism, we anticipate a reduced sensitivity in these components. In addition, the larger feature pool created by combining deformable attention with multi-scale inputs may further reduce the ablation effect. It is difficult to predict which of these opposing effects will dominate. The decoder \ac{mhsa} remains standard, as in \ac{detr}, so we anticipate no change in ablation sensitivity.
        \ac{ddetr} also increases the number of \acp{qe} from 100 to 300 and doubles their embedding size, likely leading to finer-grained representations more susceptible to ablation. The new parameters added by the increased embedding size act as learnable reference points for the first decoder \ac{mhca} layer. To isolate effects, we separate \ac{qe} and reference point ablations.
        The performance curves for \ac{ddetr}'s encoder MHSA closely resemble those of \ac{detr}, though the impact is only half as severe. We observe very similar behavior for the decoder \ac{mhca} ablations' effect on regression performance, while classification remains unaffected. With \ac{nia} we were able to show that deformable attention's reduction to the most important tokens can compensate for the disproportionate influence increase of projection matrix values due to multi-scale feature concatenation.
        Decoder MHSA again has negligible impact, aligning with the expectations.
        Ablating \ac{ddetr}'s \acp{qe} results in a sharp performance drop of \begin{math}\approx\end{math} 6\%p with only 5\% ablation, indicating greater sensitivity compared to \ac{detr}. This finding aligns with our hypothesis that \acp{qe} play a more critical role in \ac{ddetr}. However, the total performance reduction at 50\% ablation is very similar to \ac{detr}. Class-wise performance analysis of both models, reveals that the added parameters in \ac{ddetr} primarily support distinctions among the most frequent 70-80\% classes (Appendix fig. \ref{fig:detr_ddetr_cumsum_performance}).
        Ablating reference points alone results in minimal performance reduction \begin{math}\approx
        \end{math} 0.1\textperthousand, revealing that the \acp{qe} constitute the major functionality contribution (Appendix fig. \ref{fig:ddetr_query_ref}).
        The block-wise ablation results presented on the right of figure \ref{fig:detr_layers} reveal a more distributed knowledge representation compared to \ac{detr}, especially in the decoder. Due to negligible effects, the decoder \ac{mhsa} results are omitted. 
        Targeting the encoder \ac{mhsa} demonstrates that the first five blocks contribute little to the overall performance, only the final block causes a notable 3.6\%p \ac{mgiou} decrease, underscoring redundancy in earlier layers.
        The decoder MHCA is largely unaffected by single-block ablations, reinforcing the observation that deformable attention features facilitate the emergence of resilient representations.
    \subsection{Detection Transformer Model with Improved DeNoising Anchor Boxes (\ac{dino}) }
        Building on the architectural innovations introduced in the \ac{dino} paper, we formulate the following hypotheses regarding the impact of dynamic anchors and the "look forward twice" mechanism. Dynamic anchors may reduce the relevance of static content query (embeddings), by providing information tailored to each input image rather than fixed decoder inputs. The "look forward twice" mechanism introduces an additional gradient path by basing prediction boxes $b_i^{aux}$ used in auxiliary losses on predictions of the previous decoder block $b_{i-1}$ instead of its own. These boxes are corrected by their own results $\Delta b_i$. We hypothesize that this approach improves inter-block consistency in prediction across decoder blocks.
        Ablating the acp{qe} has minimal impact, as expected, because dynamic anchors directly transmit positional information from encoder to decoder. This becomes even clearer when we compare the sparsity levels, measured by the percentage of near-zero values (range +/-0.05) in the \acp{qe}: \ac{dino} shows a sparsity of 58\%, compared to just \begin{math}\approx\end{math} 4\% in \ac{detr} and \ac{ddetr}. Completely ablating the static content queries shows an increase in \ac{mgiou} by +0.03\%p and by +1.15\%p F1-score, making \ac{dino} the top performer across both metrics. This suggests that the ability that emerged from the content queries of \ac{detr} and \ac{ddetr} shifted to the encoder with \ac{dino}, making content queries obsolete.
        To validate this, we retrained \ac{dino} under standard mmdetection settings with frozen content queries set to zero. The model trained without learnable content queries achieves 81.08\% \ac{mgiou} and 86.9\% F1-score, while the model with learnable content queries obtains 81.63\% \ac{mgiou} and 85.45\% F1-score. We also completely ablated the content queries from our self-trained model to ensure that we observe the same behavior as in the mmdetection model, which we did with -0.02\%p mgIoU and +2.19\%p F1-score. Thus, the content queries seem to assist in the early stages of training, but are largely disregarded later, reflected by their increasing sparsity (67\% for our self-trained model).
        Ablations in encoder MHSA caused only minor decreases in \ac{mgiou} and F1-score.
                \begin{figure}[htb]
          \centering
           \includegraphics[width=0.8\linewidth]{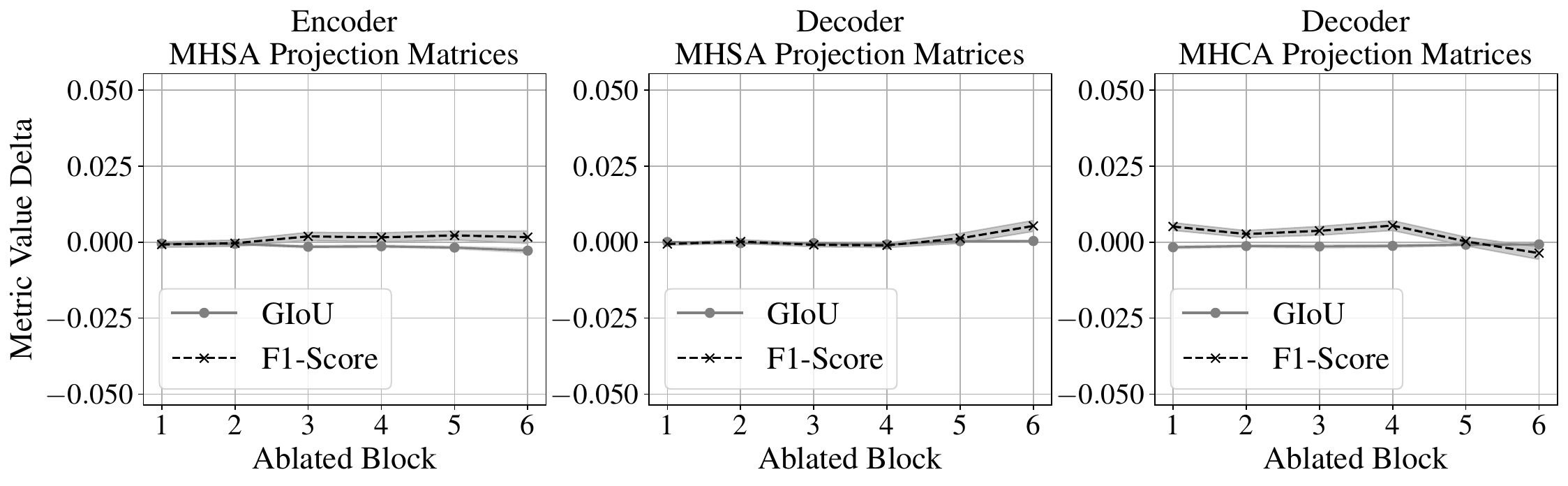}
           \caption{\ac{dino} performance difference for 30\% block-wise ablations in encoder MHSA and decoder MHCA. The shaded areas correspond to the standard deviation.}
           \label{fig:dino_layers}
        \end{figure}
        The decoder's MHSA exerts a greater influence than in \ac{detr} and \ac{ddetr}, due to its processing of dynamic anchors fused with the static queries - though this fusion may hinder convergence, given the latter's declining relevance as training progresses.
        Decoder MHCA ablations uniquely led to an increase in classification performance alongside a decrease in regression ability, suggesting a prioritization of localization over classification. This trend mirrors the observed gain in mGIoU and accompanied by slight F1-score decline from DDETR to DINO, noted at the beginning of section \ref{sec:results}.
        \ac{dino} exhibits almost no performance degradation when any attention layers are ablated block-wise as visualized in fig. \ref{fig:dino_layers}. Minor positive changes in F1-score were observed from block three onward, showing that classification benefited slightly from encoder MHSA ablations. The knowledge distribution across decoder blocks led to minor gains in classification and negligible regression loss.
        \ac{dino}’s block-wise ablations indicate that training with the “look forward twice” rule results in the aspired objective - highly distributed knowledge across blocks \cite{zhang2023dino}.

\section{Conclusion and Future Work}
    \label{sec:conclusion}
    We conducted \ac{nia} studies on \ac{detr} models trained on the \ac{coco} dataset, along the guiding question of how specific innovations introduced with \ac{detr}, \ac{ddetr}, and \ac{dino} affect the organization of their learned representations.
    Key components that differentiate \acp{detr} from \acp{cnn} and \ac{nlp} transformers—encoder and decoder MHSA, MHCA projection matrices, and \acp{qe} — were systematically ablated at varying levels (5–50\%) across all blocks and in a single block setting (30\%). To support reproducibility and further exploration, we provide the \emph{DeepDissect} library. 
    We found that the effect of ablations on model performance significantly decreases from \ac{detr} over \ac{ddetr} to \ac{dino}. \ac{detr} showed to rely heavily on the encoder \ac{mhsa} and decoder \ac{mhca} for regression and classification, while ablations in the \acp{qe} only moderately impact regression performance. However, ablating parts of the decoder MHSA has minimal impact on regression and even slightly improve classification, suggesting potential for parameter reduction and inference speed up by directly feeding \acp{qe} into the \ac{mhca} layer.  
    The combination of multi-scale features with deformable attention in \ac{ddetr} leads to a reduced effect of ablations in encoder \ac{mhca} and decoder \ac{mhca}. \acp{qe} become significantly more sensitive at 5\% ablation percentage, due to overfitting on frequent classes, caused by a parameter increase compared to \ac{detr}. The higher selectivity of the deformable attention leads to a concentration of representations important for regression in the last encoder block, where the decoder \ac{mhca} has high redundancy over all blocks.
    \ac{dino}'s look forward twice update rule leads to redundant representations for both object detection subtasks that are well distributed across the whole model. This redundancy opens up the possibility for efficiency gains via block reduction while retaining performance.
    \acp{nia} help to uncover that the direct information transfer from encoder to decoder via dynamic anchor boxes makes content queries obsolete in the fully trained state, as the function they fulfilled in \ac{detr} and \ac{ddetr} emerges in \ac{dino}'s encoder. By training \ac{dino} without static \acp{qe}, we found that they still support in the early stages of training, but become irrelevant as training progresses.
    In our future work, we aim to expand our ablation experiments to additional components like \ac{ffn} or the offset sampling and attention weight prediction in deformable attention.
    Additional experiments with the \ac{ddetr} variants utilizing refinements (one/two stage) would be interesting to see if they show comparable behavior to \ac{dino}, as these use the encoder features to enhance \acp{qe} as well. Furthermore, we would like to explore ablation target selection schemes beyond simple randomization and analyze class-level representations using specifically designed datasets in conjunction with \ac{nia} studies. Finally, we aim to establish \acp{nia} as a standard tool for improving understanding of the inner workings of deep learning models through our continued refinement of the methodology.

\newpage
\appendix
\section{Appendix}

\begingroup
\setcounter{figure}{0}
\renewcommand{\thefigure}{A\arabic{figure}}

    \begin{figure}[!htbp]
      \centering
       \includegraphics[width=0.4\linewidth]{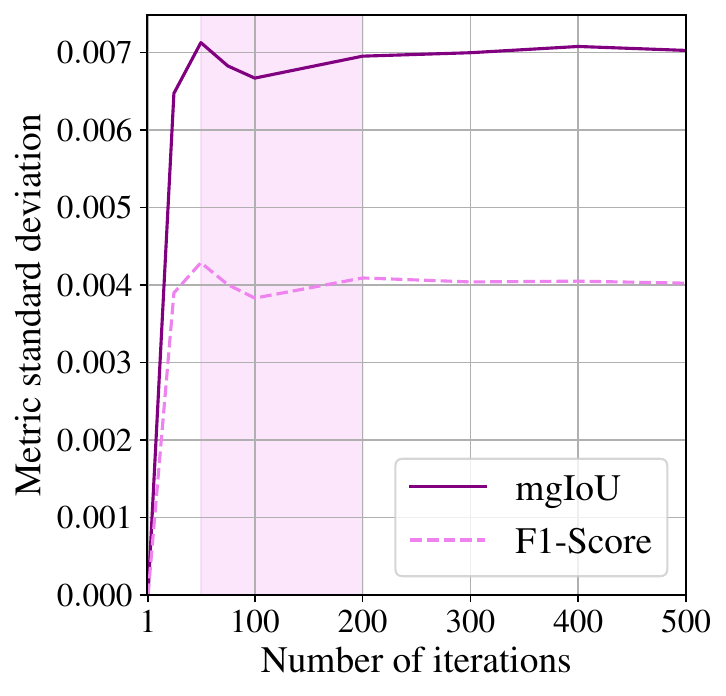}
       \caption{Standard deviation of \ac{mgiou} and F1-Score for varying numbers of random configuration in query embedding ablations.}
       \label{fig:query_embedding_std_variation}
    \end{figure}

    \begin{figure}[!htbp]
      \centering
       \includegraphics[width=1\linewidth]{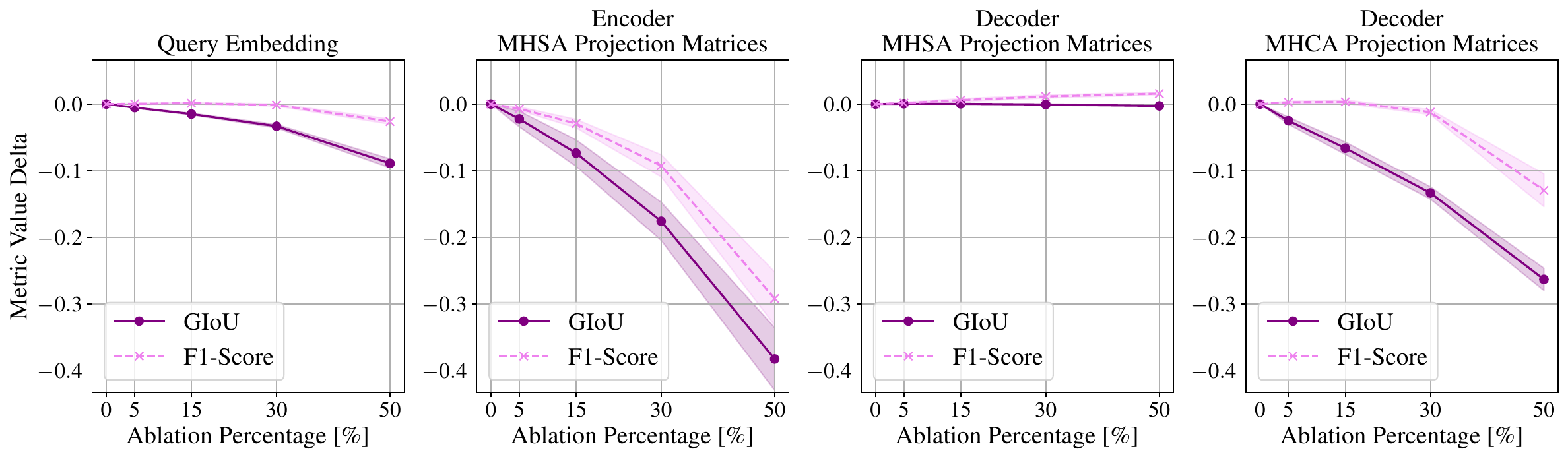}
       \caption{\ac{detr} performance differences for increasing ablation percentages in \acp{qe}, encoder MHSA, decoder MHSA and decoder MHCA. The shaded areas correspond to the standard deviation.}
       \label{fig:detr_component_ablations}
    \end{figure}

    \begin{figure}[h!]
      \centering
       \includegraphics[width=1\linewidth]{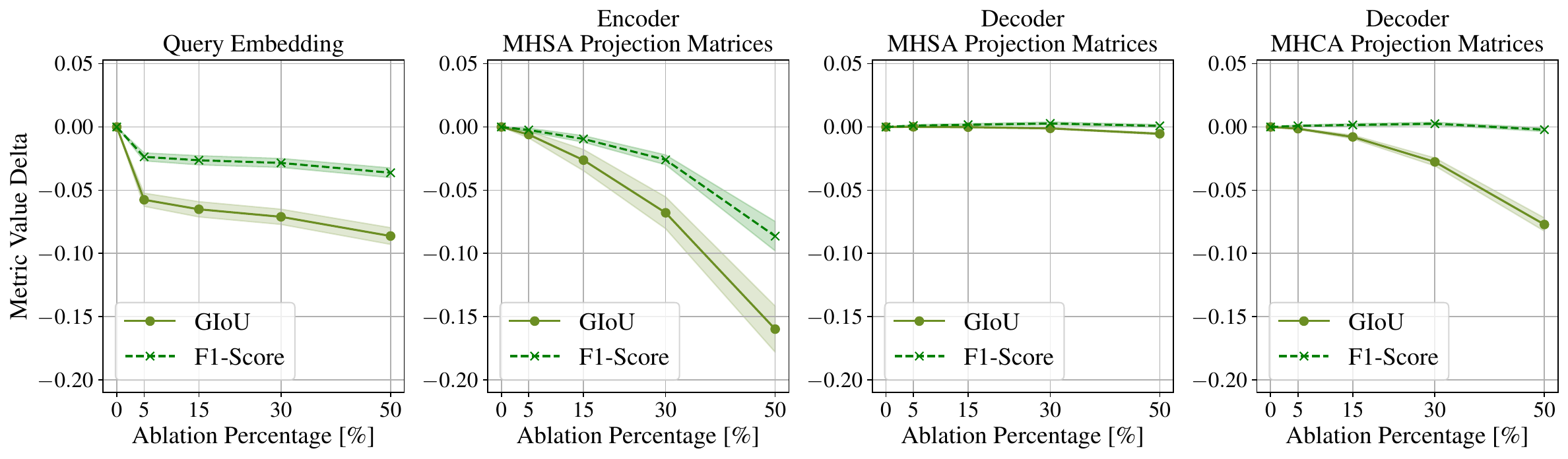}
       \caption{\ac{ddetr} performance differences for increasing ablation percentages in \acp{qe}, encoder MHSA, decoder MHSA and decoder MHCA. The shaded areas correspond to the standard deviation.}
       \label{fig:ddetr_component_ablations}
    \end{figure}

    \begin{figure}[!htbp]
      \centering
       \includegraphics[width=1\linewidth]{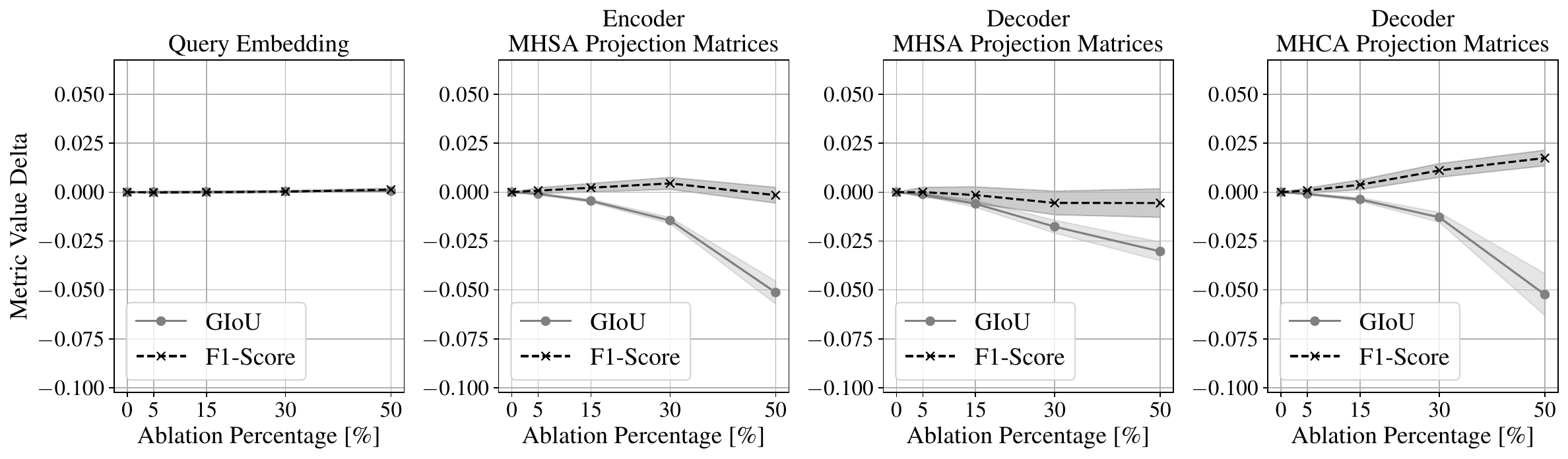}
       \caption{\ac{dino} performance differences for increasing ablation percentages in \acp{qe}, encoder MHSA, decoder MHSA and decoder MHCA. The shaded areas correspond to the standard deviation.}
       \label{fig:dino_component_ablations}
    \end{figure}

    \begin{figure}[!htbp]
      \centering
       \includegraphics[width=1\linewidth]{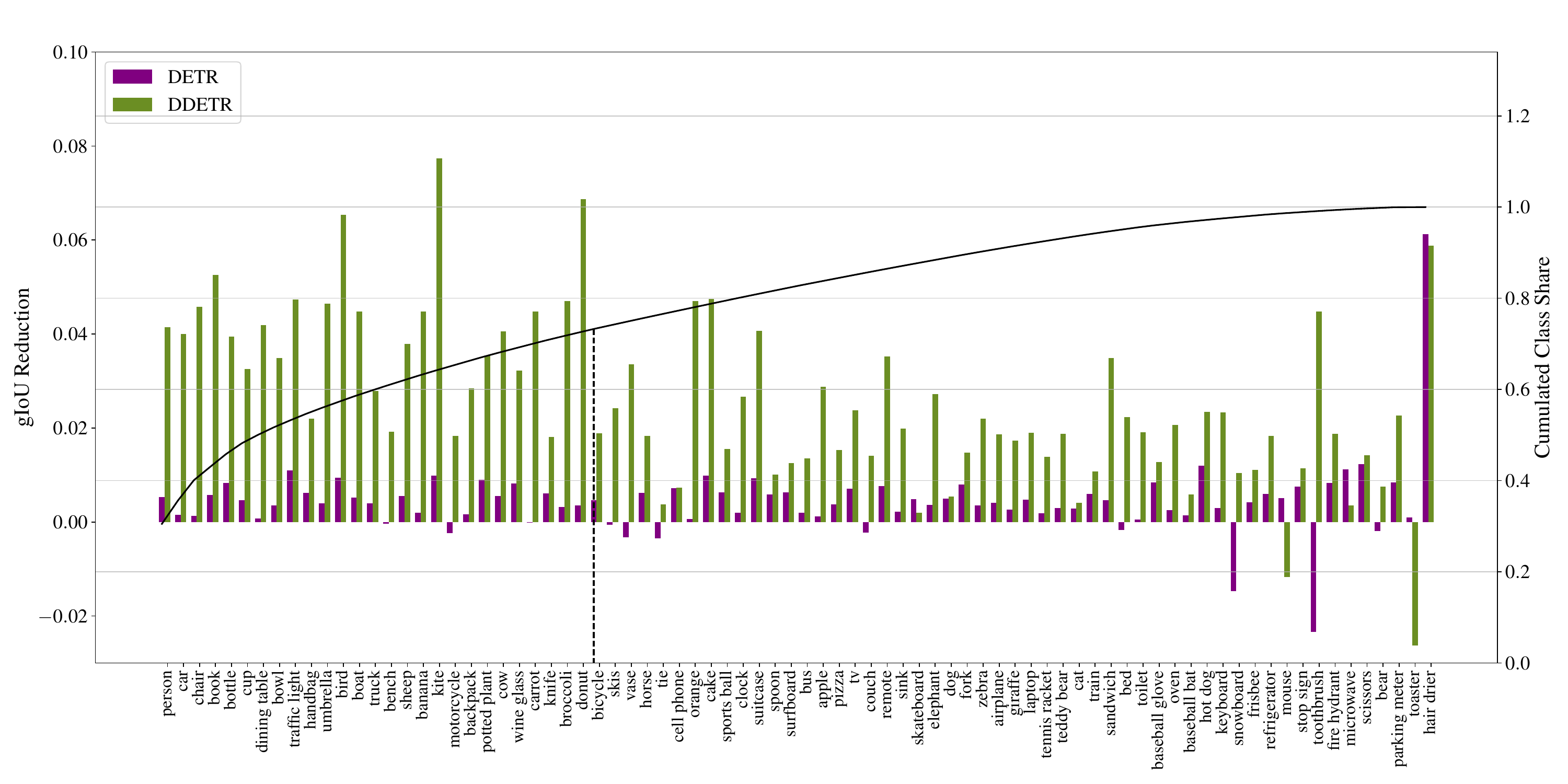}
       \caption{\ac{dino} performance differences for increasing ablation percentages in \acp{qe}, encoder MHSA, decoder MHSA and decoder MHCA. The shaded areas correspond to the standard deviation.}
       \label{fig:detr_ddetr_cumsum_performance}
    \end{figure}

    \begin{figure}[!htbp]
      \centering
       \includegraphics[width=0.7\linewidth]{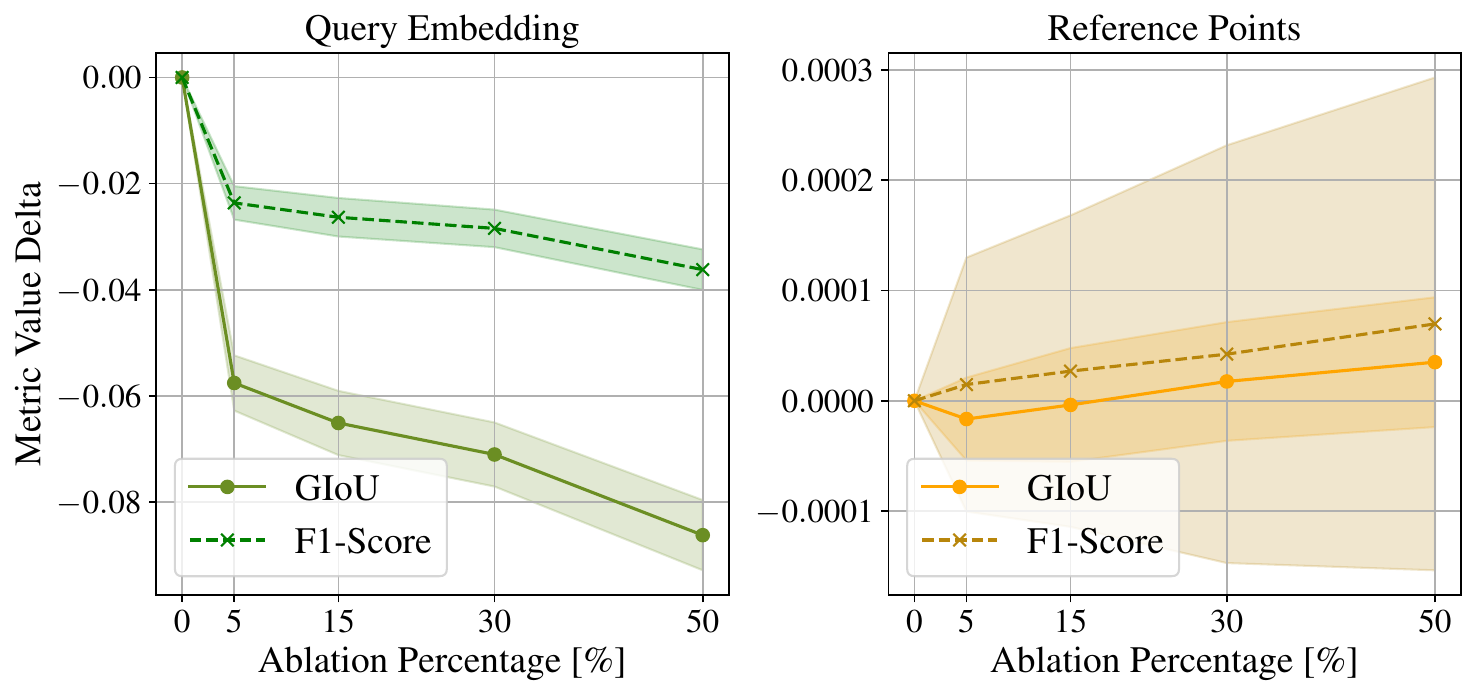}
       \caption{\ac{ddetr} performance differences for increasing ablation percentages in \acp{qe} and reference points. The shaded areas correspond to the standard deviation.}
       \label{fig:ddetr_query_ref}
    \end{figure}

\bibliographystyle{unsrt}  
\bibliography{references}

\end{document}